\def\year{2019}\relax
\documentclass[letterpaper]{article} 
\usepackage{aaai19}  
\usepackage{times}  
\usepackage{helvet}  
\usepackage{courier}  
\usepackage{url}  

\usepackage{graphicx}  
\usepackage[normalem]{ulem}
\usepackage{enumitem,setspace}
\usepackage{amsmath,amssymb,amsthm}
\usepackage{multirow}
\usepackage{makecell}
\usepackage{multirow}
\usepackage{arydshln}
\usepackage{color}
\definecolor{darkblue}{rgb}{0, 0, 0.5}

\usepackage[compact]{titlesec}
\titlespacing{\paragraph}{%
  0pt}{
  0.15\baselineskip}{
  1em}
\newcommand{\citet}[1]{\citeauthor{#1} \shortcite{#1}}

\begin{document}
\setlength{\abovedisplayskip}{4pt}
\setlength{\belowdisplayskip}{3pt}
%
\title{ScisummNet: A Large Annotated Corpus and Content-Impact Models\\ for Scientific Paper Summarization with Citation Networks}

\author{{\large Michihiro Yasunaga \quad\quad Jungo Kasai \quad\quad Rui Zhang}\\[1mm]{\bf {\large Alexander R. Fabbri \quad Irene Li \quad Dan Friedman \quad
Dragomir R. Radev}}\\[1mm]
Department of Computer Science, Yale University\\
\scalebox{0.85}[0.9]{{\tt \{michihiro.yasunaga,r.zhang,alexander.fabbri,irene.li,dan.friedman,dragomir.radev\}@yale.edu}}\\
\scalebox{0.85}[0.9]{{\tt jkasai@cs.washington.edu}}
}
\maketitle

\begin{abstract}
\vspace{-1mm}
Scientific article summarization is challenging:
large, annotated corpora are not available,
and
the summary should ideally
include
the article's impacts
on research community.
This paper provides novel
solutions to these two challenges.
We
1)
develop and release the first large-scale manually-annotated corpus for scientific papers (on computational linguistics) by enabling faster annotation,
and
2)
propose summarization methods that integrate the authors' original highlights (abstract) and the article's actual impacts on the community (citations), to create comprehensive, hybrid summaries.
We conduct experiments to demonstrate the efficacy of our corpus in training data-driven models for scientific paper summarization and the advantage of our hybrid summaries over abstracts and traditional citation-based summaries.
Our large annotated corpus and hybrid methods provide a new framework for scientific paper summarization research.\footnote{Our dataset is available at \url{https://michiyasunaga.github.io/projects/scisumm_net/}}
\vspace{-2mm}
\end{abstract}

\section{Introduction}
Fast-paced publications in scientific domains motivate us to develop
automatic summarizers for
scientific articles.
Recent work in automatic summarization has achieved remarkable performance for news articles:
Single-Document Summarization \cite{parveen-ramsl-strube:2015:EMNLP,cheng-lapata:2016:P16-1,see2017get,narayan2018ranking}, Multi-Document Summarization \cite{hong-nenkova:2014:EACL,cao2015ranking,cao2017improving}.
Scientific article summarization, on the other hand,
is less explored, and differs from news article or other general summarization.
For example, scientific papers are typically longer and contain more complex concepts and technical terms.
Moreover, they are structured by section and contain citations.

\begin{figure}[!t]
    \vspace{-2mm}
    \hspace{-2mm}
    \centering
    \includegraphics[width=0.48\textwidth]{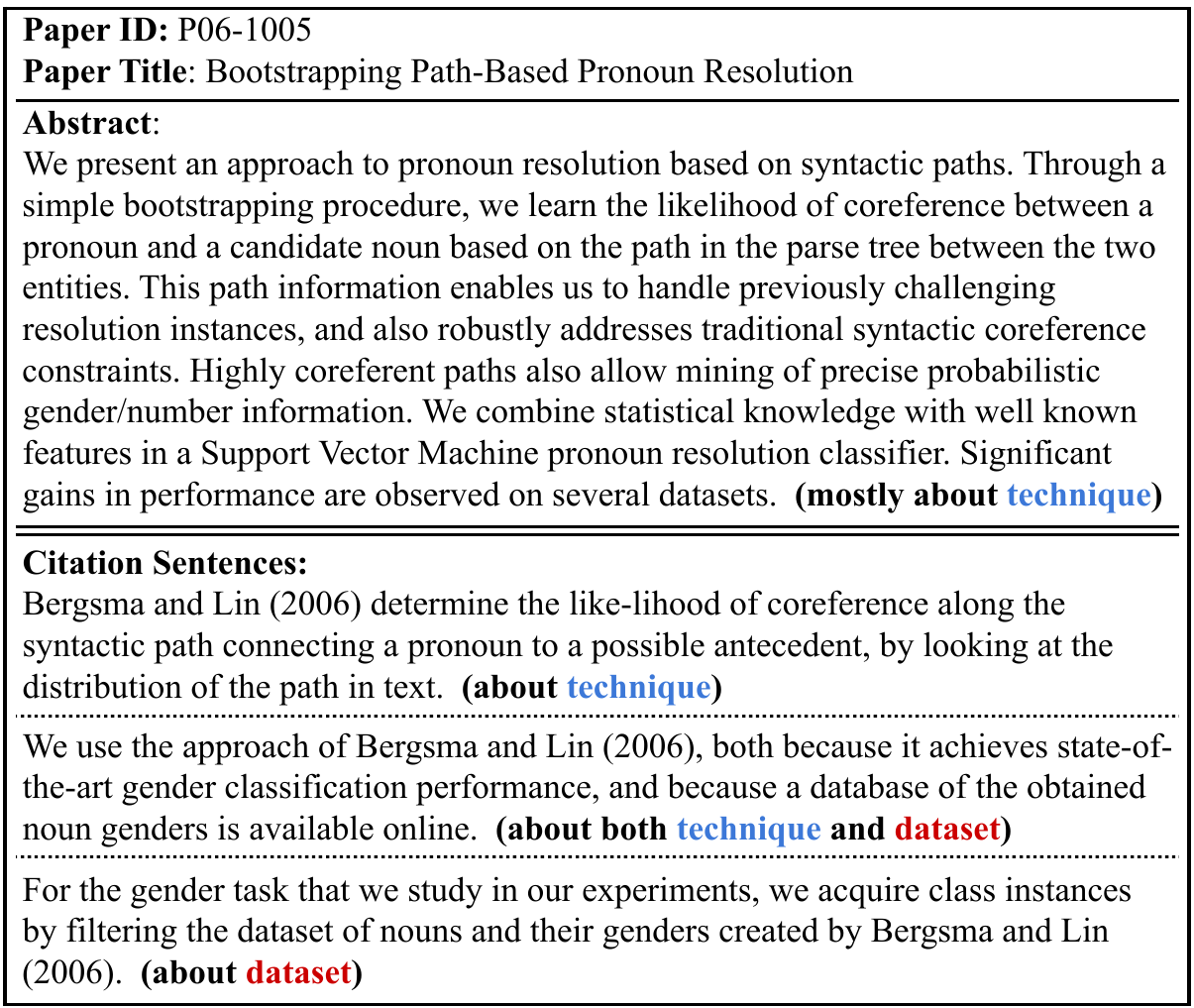}\vspace{-3mm}
    \caption{
    Abstract and citations of \protect\cite{bergsma-lin:2006:COLACL}. The abstract emphasizes their pronoun resolution techniques and improved performance; the citation sentences reveal that their noun gender dataset is also a major contribution to the research community, but it is not covered in the abstract.
    }
\label{fig:citation}
\vspace{-4mm}
\end{figure}

\urlstyle{same}
To encourage research in scientific article summarization, several shared tasks have been organized recently: TAC 2014 (biomedical domain),
CL-SciSumm \cite{jaidka2016overview,jaidka2017overview,jaidka2018overview} (computational linguistics domain, consisting of ACL Anthogoly papers).
While these shared tasks have established a foundation for scientific paper summarization, their datasets are small, with just 
30-50 articles.
As understanding and annotating a scientific paper require domain-specific expert knowledge, annotation does not scale to a large corpus as compared to news articles, preventing us from
applying data-driven approaches such as neural networks shown powerful in news article summarization
\cite{cheng-lapata:2016:P16-1,see2017get}.
In news article summarization, on the other hand,
prior work \cite{woodsend-lapata:2010:ACL,cheng-lapata:2016:P16-1}
has manually
created gold summaries for 9,000 documents
and extended them to 200K documents by heuristics.
This type of annotation or crowd-sourcing is not realistic for scientific papers due to their length and technical content.

Another characteristic of scientific papers is that they may have impacts that are not expected at the time of publication.
For instance (Figure \ref{fig:citation}), the abstract of \citeauthor{bergsma-lin:2006:COLACL} emphasizes their techniques and
improved
performance in pronoun resolution, but a citation analysis reveals that their contribution to subsequent work lies largely in the noun gender dataset they created.
While the abstract of a paper provides a solid summary of the {\it content} from the authors' point of view, it may fail to convey the actual {\it impact} of the paper on the research community.
Additionally, the significance of a paper may change over time
due to the progress and evolution of
research \cite{mei-zhai:2008:ACLMain}.
In such situations our summary should ideally accommodate not only the major points highlighted by the authors (abstract) but also the views offered by the scientific community (citations).

This paper presents a novel dataset and summarization method to tackle the aforementioned problems in scientific paper summarization.
Our corpus, which contains the citation network of ACL Anthology papers and human-written summaries for the 1,000 most cited papers, expands the existing CL-SciSumm project \cite{jaidka2016overview} and provides the largest manually-annotated
dataset for scientific paper summarization.
For each of the 1,000 papers (we call {\it reference papers}, or {\it RPs}), experts in CL/NLP read its abstract and incoming citation sentences to create a gold summary.
This way, annotators can grasp broad, major aspects of the RP
without reading the whole text,
enabling faster annotation.
We also conduct studies to validate that summaries created in this method are actually as comprehensive as summaries created by reading the full papers.
Our dataset (1,000 papers) is
significantly larger than the prior CL-SciSumm corpus (30 papers) and serves as a useful resource for supervised scientific paper summarization.

Further, we propose two novel summarization models for scientific papers that
capture
both the papers' \textit{content} highlighted by the authors and \textit{impact}
perceived by
the research community
(hybrid summarization).
In both
models, given a reference paper (RP) to summarize, we take its abstract as the authors' insight, and identify a set of text spans ({\it cited text spans}) in the RP that are referred to by incoming citation sentences (i.e., community's views).
The first approach then
summarizes the union of the abstract and cited text spans, to integrate both components.
The second approach, motivated by the fact that we already have the abstract as a clean self-summary of the paper, augments the abstract by adding salient texts extracted from the cited text spans (i.e., the community's views not covered in the abstract).
For both approaches we also exploit the citation counts of the RP and its citing papers as an additional feature, to better reflect the authority of each work in the research community.
To experiment with these two methods, we implement two neural network-based summarization models, which are also motivated by the architecture of \citet{yasunagaetal2017}'s neural multi-document summarizer.

In evaluation, we use the CL-SciSumm shared task \cite{jaidka2016overview}, an established benchmark for scientific paper summarization.
This benchmark dataset contains gold summaries that are created by experts who read papers and their citation sentences.
First, we find that
our large training corpus enables neural summarizers to boost their performance and outperform all prior participants in the shared task. This confirms the usefulness of the proposed dataset.
Second, we demonstrate that the proposed hybrid summarization methods can indeed incorporate both the authors' and research community's views, thereby producing more comprehensive summaries than abstracts.
In summary, our contributions are as follows.
\vspace{-1mm}
\begin{itemize}
\setlength{\itemsep}{-0.5mm}
    \item A large manually-annotated corpus (1,000 examples) for scientific article summarization that facilitates research on supervised approaches.
    \item Novel scientific paper summarization methods that integrate both the authors' and research community's insights (hybrid summarization).\vspace{-1mm}
\end{itemize}

\section{Background \& Motivation}
\subsection{Text Summarization}
Many existing summarization systems employ extractive methods to produce a summary, typically by ranking the salience of each sentence in a given document and then selecting sentences to be included in the summary \cite{erkan2004lexrank,parveen-ramsl-strube:2015:EMNLP}.
Recently, in news article summarization, neural network-based approaches have proven successful \cite{cao2015ranking,cheng-lapata:2016:P16-1,nallapati2017summarunner,see2017get}.
This work presents neural network-based extractive models for scientific paper summarization.

\subsection{Scientific Paper Summarization}
\label{sec:motivation_scisumm}

Scientific paper summarization has been studied for decades
\cite{Paice:1980:AGL:636669.636680,Elkiss:2008:BME:1331122.1331127,lloret2013compendium,jaidka2016overview,parveen2016generating}.
While early work \cite{luhn1958automatic,Paice:1980:AGL:636669.636680,Paice:1993:IIC:160688.160696} focused on producing \textbf{content-based summaries} of target papers,
the use of citations was later proposed to summarize
target papers' contributions and lasting influence on the research community.

\begin{figure*}[!th]
    \vspace{0mm}\hspace{-1.5mm}
    \centering
    \includegraphics[width=0.83\textwidth]{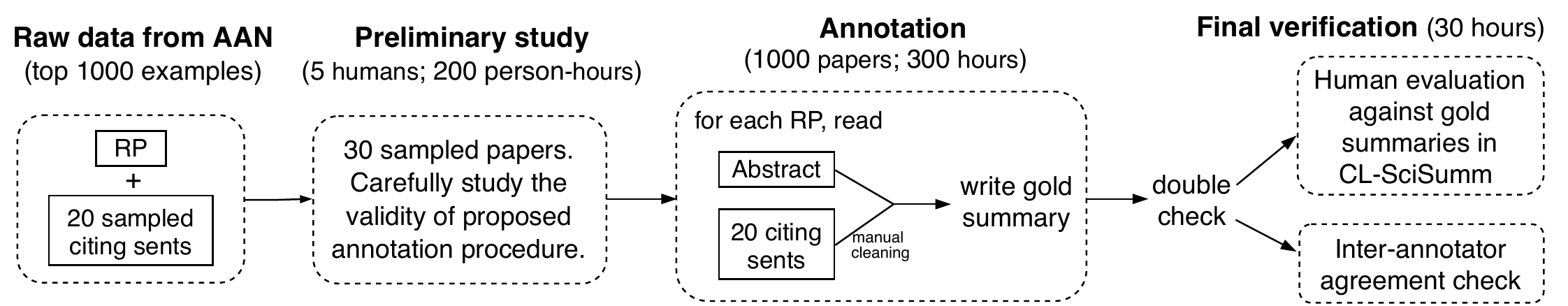}\vspace{-2mm}
    \caption{Overview of the dataset construction process.
    }
\label{fig:data_overview}
\vspace{-3mm}
\end{figure*}

\paragraph{Citation-based summarization.}
Early work in citation-based summarization
\cite{nakov2004citances,Elkiss:2008:BME:1331122.1331127,qazvinian2008scientific,abuJbara&Radev11a} aimed to summarize the contribution of a target paper (often called {\it reference paper}, or {\it RP}, in this context)
by extracting a set of sentences from the citation sentences.
We call a sentence that cites the RP a {\it citation sentence} (or
{\it citing} sentence).
A citation sentence can be viewed as a short summary of the RP written from the citing authors' perspective.
Hence, a collection of citation sentences reflects the impact of the RP on the research community \cite{Elkiss:2008:BME:1331122.1331127}.
While citation sentences provide
the community's views of
the RP, prior work \cite{siddharthan2007whose,mei-zhai:2008:ACLMain} pointed out issues in using citation sentences directly for summarization.
In citing sentences, the discussion of the RP is often mixed with the content of the citing paper or with the discussion of other papers cited jointly, containing much irrelevant information.

To address such issues, recent work
\cite{mei-zhai:2008:ACLMain,cohan2015sci,jaidka2016overview,li2017computational,Cohan:2017:CCS:3077136.3080740,cohan2017scientific}
considers {\it cited text spans}-based summarization, where
they identify a {set} of text spans ({\it cited text spans}; often a set of sentences) in the RP that
its citing sentences refer to, and
perform summarization on
the identified text spans.
This way, while the summary consists of words in the RP, it reflects the research community's insights.
Experimental results in \citeauthor{mei-zhai:2008:ACLMain} show that their cited text span-based model outperforms direct summarization of citing sentences.
Cited text span-based summarization is also adopted as the default approach in
two recent shared tasks on scientific paper summarization: TAC 2014
and CL-SciSumm
\cite{jaidka2016overview}.
Their datasets provide each RP and its incoming citation sentences, cited text spans, and a gold summary written by experts; the participants are asked to produce summaries using the RP and its citation sentences.

\paragraph{Our hybrid models.}

While the aforementioned citation-based summarization techniques inform us of the impact of an RP, they may overlook the authors' original message.
For example, citation sentences (and consequently, cited text spans) often focus on the conclusion of the RP and may not cover other important aspects such as the motivation of the work.
Moreover, in our preliminary study conducted on the CL-SciSumm shared task, we found that the quality of cited text span-based
summaries produced by participants, often falls short of the abstracts in ROUGE evaluation against gold summaries.
\citet{conroy2017section} also find that the terms from abstracts in scientific documents often
cover a large portion of human summaries.
Our motivation in this work is therefore to integrate both the authors' original highlights (abstract)  and research community's views (citations),
and ultimately to improve upon the abstract.

\paragraph{Datasets.}
Previous datasets for scientific document summarization
are small (\citeauthor{Teufel:2002:SSA:638178.638180,jaidka2016overview}; TAC 2014), with only several dozen articles.
Consequently, most of the existing summarizers for scientific papers are
unsupervised or tuned on small data \cite{abuJbara&Radev11a,cohan2015sci,clscisumm_sys8}.
In fact, in the previous CL-SciSumm shared task (30 data examples), no data-driven approaches
like
neural networks saw great success.
The new dataset we introduce
here
(1,000 examples) is much larger than the prior CL-SciSumm corpus, enabling data-driven approaches to scientific paper summarization.
In our experiments,
we show that our dataset indeed allows neural network-based summarization models to
outperform all prior participants in the shared task.

Recent work by \citet{collins2017supervised} and \citet{cohan2018discourse} is related to ours in that they also introduce large-scale datasets and neural summarization models for scientific papers. Yet,
while they focus on content-based summarization with automatically created gold summaries, our work constructs manually-annotated gold summaries as well as citation information to study the research community's view on each reference paper.



\section{Dataset Construction}
\label{data_construction}

To overcome data scarcity in scientific paper summarization,
we develop and release a manually-annotated, large-scale corpus for research papers in computational linguistics (CL).
Our corpus contains the 1,000 most cited papers in the ACL Anthology Network (AAN) \cite{aan_2013},
their citation information, and gold summaries annotated by experts in the field.
We follow the format of two prior datasets of scientific paper summarization, CL-SciSumm \cite{jaidka2016overview} and TAC 2014 (biomedical domain),
so that systems trained / tested on our corpus can also be applied to or evaluated on those established datasets.
Figure \ref{fig:data_overview} depicts our data construction process.

\subsection{Data Processing}
We extract the 1,000 most cited papers and their citation sentences from
AAN.
The 1,000 papers have 21\,-\,928 citations in the anthology.
For each of the RPs, we sample and clean 20 citation sentences, which are usually sufficient to study the research community's views of the RP \cite{mei-zhai:2008:ACLMain}.
Specifically, following the prior datasets, we keep the oldest and latest citations and randomly sample the rest so that the 20 citations cover an extended period of time.
We then remove inappropriate citation sentences (i.e., list citations, tables, those with bugs) and clean the rest, resulting in 15 citation sentences on average for each RP.

\begin{figure*}[!t]
    \vspace{0mm}\hspace{-1.5mm}
    \centering
    \includegraphics[width=0.77\textwidth]{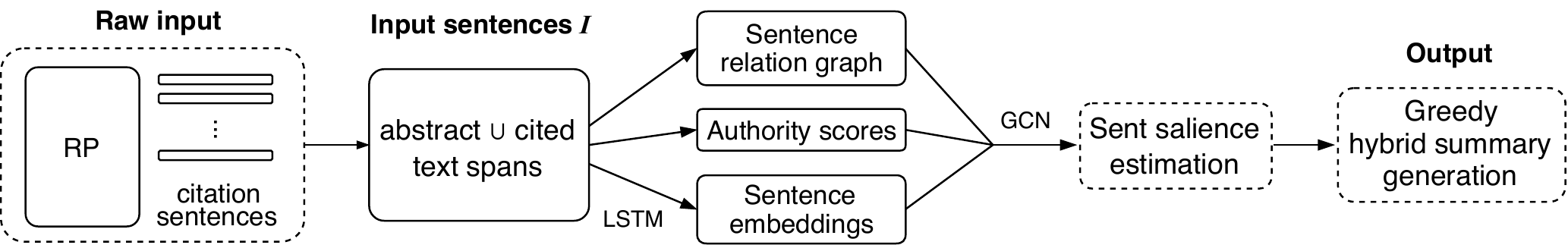}\vspace{-2mm}
    \caption{Overview of our summarization models.
    }
\label{fig:model_overview}
\vspace{-3mm}
\end{figure*}

\subsection{Annotation}

We aim to develop gold summaries for the 1,000 papers in CL.
In the prior datasets (CL-SciSumm and TAC 2014, containing $\sim$30 papers), gold summaries were prepared by humans with domain expertise, in the following manner
(i.e., {\it expert} summaries):
given a RP to summarize, the annotators
read all the text of the RP and its citation sentences to grasp its content and impact, and wrote a comprehensive summary.
Yet, due to the length and technical content of scientific papers, such annotation requires a significant amount of time as well as expertise, hindering the construction of a large-scale corpus for scientific article summarization.

To scale our annotation to the 1,000 papers, we develop a faster annotation procedure in this work.
Five PhD students in NLP or people with equivalent expertise divide the 1,000 RPs, and read each paper's abstract and incoming citation sentences.
Then, he or she identifies a few
salient citation sentences that convey the RP's specific contributions \uline{not covered in the abstract} and
write a gold summary based on the abstract and selected citation sentences.
This way, the annotators can save the time of reading the whole text of the RP, but can still grasp broad aspects of the paper to create a comprehensive summary.
To take
\citeauthor{bergsma-lin:2006:COLACL} (Figure \ref{fig:citation})
as an example again,
while its abstract elaborates on the pronoun resolution techniques,
its citation sentences reveal other major aspects of the RP such as the contribution of a noun gender dataset, which is discussed in some part of the RP but is not highlighted in the abstract.
An expert summary would include both of these aspects to describe the RP.
By reading the citation sentences in addition to the abstract, we can comprehend much of the content and impact of the RP without reading all its text.

Lastly, the citation sentence cleaning / selection process is double-checked to prevent mistakes.

\paragraph{Validation of our annotation procedure.}

Prior to
annotation, we conducted preliminary studies on
30 sample papers.
For each, we had the annotators list out the
summary-worthy points they found 1) by reading the abstract + citing sentences, and 2) by reading the full paper; we observed that on average
the former (our annotation method)
covered
over 90\% of the major points found by reading the full papers, while just requiring 30\% annotation time.
This study suggests that by reading the abstract + citing sentences, annotators can create summaries comparable in quality to the ground truth in an inexpensive way.

\paragraph{Statistics.}
Our annotated summaries resulted in 151 words on average (similar to the gold summaries in the CL-SciSumm corpus, 150 words).
To study the inter-annotator agreement on determining salient citations, we randomly picked 40 RPs
and assigned
another annotator; the Kappa coefficient \cite{cohen1960coefficient}
for inter-annotator agreement was 0.75, {\it substantial agreement}, on \citeauthor{landis1977measurement} scale.
The high inter-annotator agreement
further supports the
efficacy of using abstract + citing sentences to create summaries.

\paragraph{Human evaluation.}

We also conducted human evaluation of
our gold summaries against the gold summaries
in
the CL-SciSumm corpus,
which were created by reading full papers.
We studied the 15 papers in our corpus that already exist in CL-SciSumm.
For each paper,
we asked 5 computer science students who took an NLP course
to evaluate which gold summary (ours or CL-SciSumm's) is more comprehensive, on an integer scale -2 to 2: 2 if the former (ours) is more comprehensive; -2 if the latter; 0 if they are similar; and 1, -1 are in between.
The evaluated scores were
54\% zero, 22\%
positive, and 20\%
negative, with average
+0.02.
This result indicates that our annotated summaries are comprehensive, and comparable to
or slightly better than
the summaries created by reading full papers.\vspace{0.5mm}

The dataset construction took
600+ person-hours.
This large corpus
can be used to train scientific paper summarization models that utilize citations, facilitating research in supervised methods.
In the next sections, we introduce data-driven hybrid summarization models and experiment on the proposed corpus.


\section{Hybrid Summarization Models}

Given a reference paper (RP) and its incoming citation sentences, our hybrid summarization models aim to reflect both the authors' and research community's voices on the RP.
Specifically, we regard the abstract of the RP as the authors' original perspective, and obtain the research community's insights by identifying cited text spans in the RP (i.e., sentences in the PR that are referred to by the citation sentences).
In this work, we consider the following two versions of hybrid summarization.
\begin{itemize}[topsep=5pt]
    \setlength{\leftskip}{14.1mm}
    \setlength{\parskip}{-0.5mm}
    \item[{\bf \scalebox{1}[1]{Hybrid 1:}}] Summarizing the combination of the abstract and cited text spans
    \item[{\bf \scalebox{1}[1]{Hybrid 2:}}] Augmenting the abstract with salient texts extracted from cited text spans
\end{itemize}\vspace{-1mm}
The motivation of Hybrid 2 is to build upon the clean self-summary provided by the authors and to add the community's views not covered in it.
In both models, we take the union of the abstract and cited text spans as input $I$ for summarization.
Note that the input sentences in $I$ (in particular, cited text spans) are not necessarily contiguous in the RP.
This situation is analogous to
multi-document summarization (MDS),
which aims to
produce a summary for a set of separate documents.
Motivated by graph-based MDS methods \cite{erkan2004lexrank,yasunagaetal2017},
we build a graph capturing the relations among the input sentences in $I$,
and apply a Graph Convolutional Network (GCN) \cite{kipf2017semi} on top
to perform
summarization.

\subsection{Pre-processing}
Given a reference paper (RP) and its incoming citation sentences, we first prepare the input sentences for summarization (i.e., abstract $\cup$ cited text spans), and build their sentence relation graph.

\paragraph{Cited text spans.}

We extract cited text spans in the RP for each incoming citation sentence, and then compile them for all the given citations.
To identify cited text spans for a given citation sentence, we choose top two sentences in the RP that are most similar in terms of the tf-idf cosine similarity measure (stop words excluded).

We repeat the extraction for all the given citation sentences, and take the union to construct the complete cited text spans of the RP.
The union of the abstract and cited text spans of the RP will be the input $I$ for summarization.
In our experiments $I$ contained about 40 sentences on average.

\paragraph{Sentence relation graph.}

We build a graph
that takes the input sentences as nodes and captures their relationships via edges.
We adopt the widely-used
cosine similarity graph \cite{erkan2004lexrank}, where
every pair of sentences has an edge with a weight equal to their tf-idf cosine similarity.

\paragraph{Authority feature.}

While cited text spans provide insights by the research community, they do not necessarily reflect the authority of each citation.
\citet{mei-zhai:2008:ACLMain} argue that a citation made by a highly authoritative paper should be weighted more than that made by a less authoritative paper.
To better reflect the authority in the research community,
we consider an extra feature (authority score) for each cited text span, which is the sum of its citing papers' citation counts.
Sentences in the abstract are given the citation count of the RP.
We obtain
citation counts from the ACL Anthology Network \cite{aan_2013}.

\subsection{Main Architecture}

Given the input sentences and their relation graph,
we apply a GCN \cite{kipf2017semi}
to encode the whole input text together with the graph and to estimate the salience of each sentence in the global context.
Based on the salience scores, Hybrid 1 and 2 employ two greedy heuristics to select sentences to be included in the summaries.

\paragraph{Graph convolutional network (GCN).}
GCNs are neural networks that operate on graphs
to induce node features
based on graph structure.
GCNs have been shown effective not only in node classification tasks \cite{kipf2017semi}, but also in NLP applications such as
syntactic tree-based
sentence encoding \cite{marcheggiani2017encoding}.

Given a graph $G$ with $N$ nodes,
a GCN takes
\begin{itemize}[topsep=5pt]
    \setlength{\parskip}{-1mm}
    \item $\tilde{A} \in \mathbb{R}^{N \times N}$,\vspace{-0.3mm} the adjacency matrix of graph $G$ with added self-connections.
    \item $X \in \mathbb{R}^{N \times D}$,\vspace{-0.3mm} the input node features ($D$ is the dimension of the feature vector for each node).
\end{itemize}\vspace{-1mm}
and outputs high-level node features, $Z \in \mathbb{R}^{N \times D}$, which encode the graph structure.
The function takes a form of layer-wise propagation.
Specifically,
in an $L$-layer GCN, the propagation from the $l$-th layer to the $(l\!+\!1)$-th layer is:
\begin{equation}
\label{eq:gcn}
    H^{(l+1)} = \sigma\left(\tilde{D}^{-\frac{1}{2}}\tilde{A}\tilde{D}^{-\frac{1}{2}}H^{(l)}W^{(l)}\right)
\end{equation}
where $H^{(l)}\in \mathbb{R}^{N \times D}$ denotes the $l$-th hidden layer, with $H^{(0)}=X, H^{(L)}=Z$.
The adjacency matrix $\tilde{A}$ is normalized via the degree matrix $\tilde{D}$.
$\sigma$ is an activation function such as $\rm{tanh}$.
$W^{(l)}$ is the learnable parameter in the $l$-{\text{th}} layer.

\paragraph{Sentence encoding.}
Given the input sentences $\{s_1,s_2,$ $\dots, s_N\}$ in $I$ and their relation graph $G$, we first encode each sentence $s_i$ by applying a
Long Short-Term Memory (LSTM) \cite{Hochreiter:1997:LSM:1246443.1246450}
on its word embeddings, and taking the final
state of the LSTM as its initial sentence embedding, $\mathbf{x}_i\in \mathbb{R}^{D-1}$.
The authority score of
sentence $s_i$ can be appended to $\mathbf{x}_i$ as an additional feature.
The
sentence embeddings $\mathbf{x}_i \in\mathbb{R}^{D}$ \scalebox{0.9}{$(i\!=\!1,2,\dots,N)$} are then grouped as
a node feature matrix $X \in \mathbb{R}^{N\times D}$,
and
fed into a GCN
with the adjacency matrix $\tilde{A}$ of the sentence relation graph $G$.
Through multiple layers of propagation, the GCN
encodes the whole input text and induces higher-level sentence embeddings based on the structure of $G$.
The output of the GCN, $Z\in\mathbb{R}^{N\times D}$, gives updated sentence embeddings $\mathbf{s}_i \in\mathbb{R}^{D}$ that incorporate the global context.

\paragraph{Salience estimation.}

For each sentence $s_i$ in our input, we estimate its salience score as follows:
\begin{equation}
\label{eq:score-softmax}
  \hat{R}(s_i) = \frac{\exp (\mathbf{v}^T\mathbf{s}_i)}{\sum_{s_j\in I} \exp(\mathbf{v}^T\mathbf{s}_j)}
\end{equation}
where $\mathbf{s}_i$ is the updated embedding of sentence $s_i$, and $\mathbf{v}$ is a learnable parameter for projecting embeddings to be scalar scores.
Note that the salience scores are normalized via softmax to be a probability distribution over all the input sentences.

\subsection{Training}
The model parameters include the weights in the LSTM and GCN, and
$\mathbf{v}$.
The model is trained to minimize the cross-entropy loss between the target salience scores (true labels) $R$ and the estimated salience scores $\hat{R}$ of the input sentences:
\begin{equation}
\label{eq:loss}
L = -\sum_{s_i \in I} R(s_i)\log(\hat{R}(s_i))
\end{equation}
To construct the target scores $R$, we first take the average of ROUGE-1 \& 2 scores for each sentence $s_i$ evaluated with the gold summary \cite{cao2015ranking}, and then
rescale the
scores as a probability distribution over all the input sentences.

\subsection{Summary Generation}
Based on the salience scores estimated for the input sentences, Hybrid 1 and 2 employ two greedy heuristics to select sentences for the summaries.

\paragraph{Hybrid 1 ({\em extractive summarization of abstract \scalebox{0.95}[1]{$\cup$} cited text spans}).}
First, we sort all sentences in $I$ in descending order of the salience score.
We dequeue one sentence from the list and append it to the current summary if the sentence is of a reasonable length (more than 8 words, as in \cite{erkan2004lexrank}) and is non-redundant.
A sentence is
redundant if it is
similar to any sentence already in the summary, with tf-idf cosine similarity above 0.5 \cite{hong-nenkova:2014:EACL}.
We keep adding sentences to the summary in this way
until we reach the length limit.
Finally, sentences in the summary are sorted in the original order in the RP.

\paragraph{Hybrid 2 ({\em augmentation of abstract with salient cited text spans}).}

We take all the cited text spans from $I$ and sort them in descending order of the salience scores.
Starting from the full abstract as the initial summary, we deque one sentence from the list of cited text spans and add to the current summary if it is of a reasonable length and is non-redundant.
We repeat until the length limit, and finally sort the summary sentences in the original order in the RP.


\section{Experiments}

We experiment
the hybrid summarization models on our training corpus to study the efficacy of the proposed dataset and models.
We aim to
show that our large-scale corpus allows the data-driven neural
models to outperform prior work.
We also analyze the outputs of hybrid summarization and illustrate their advantage over abstracts and traditional citation-based summaries.

\subsection{Datasets \& Evaluation}
We train the GCN summarization models on our proposed corpus with 1,000 examples of RPs, citation sentences, and gold summaries.
{All models are validated and tested on established benchmarks},
CL-SciSumm 2016 dev \!/\! test, where the gold summaries were created by experts reading full papers.
In training, we exclude the few RPs in our corpus that also appear in the validation or test set.

We evaluate system summaries against the gold summaries
by ROUGE \cite{lin2004rouge}, which serves as a good metric for this work, as we aim to measure the comprehensiveness of summaries.
To ensure comparability with the CL-SciSumm shared task, we measure ROUGE-2 Recall, F1 (2-R, 2-F) and -SU4 F1
(SU4-F), with the same configurations:
-n 4 -2 -4 -u -m -s -f A.

\subsection{Experimental Design}

\noindent We conduct the following two experiments to study the proposed 1) corpus and 2) hybrid methods.
\paragraph{Exp 1.}

First, we study the usefulness of our dataset for data-driven models, by comparing the model performance after training on our corpus and after training on the existing CL-SciSumm corpus.
For data-driven systems, we experiment with our GCN model.
As the participants in the CL-SciSumm shared task adopted cited text span-based summarization,
to ensure a fair comparison,
we also
let these models just summarize cited text spans.
Specifically, we just select cited text spans, given the predicted salience scores
(we call this {\it GCN Cited text spans}).
We follow the same protocol as the shared task (no authority feature; summary length 250 words).

\paragraph{Exp 2.}
Next,
we study the efficacy of the hybrid summarization models.
As our goal is to learn to produce the gold summaries
(average length 150 words)
and 
compare them with abstracts\footnote{The average length of abstracts is 110 words.} or
traditional citation-based summaries,
we experiment with the GCN Hybrid models with summary length 150 words (with \!/\! without authority feature), and analyze the output hybrid summaries against those baselines.

\subsection{Training Details}

We use 100-dimensional word embeddings for the input to the LSTM sentence encoder.
The word embeddings are initialized with
GloVe \cite{pennington-socher-manning:2014:EMNLP2014}.
We set the dimension of the LSTM \!/\! GCN hidden states to be 200, 201 (i.e., \scalebox{1}{$D\!=\!201$}), and use two hidden layers for the GCN (i.e., \scalebox{1}{$L\!=\!2$}).
We apply dropout \cite{JMLR:v15:srivastava14a} to the input word embeddings as well as the outputs of the LSTM and GCN, with dropout rate 0.5.

The model parameters and word embeddings are trained by the Adam optimizer \cite{kingma2015adam}, with batch size 5, learning rate 0.001, and a gradient clipping of 2.0 \cite{Pascanu2012}.
We employ early stopping \cite{caruana2001overfitting} based on the validation loss to prevent overfitting.

\subsection{Results \& Discussion}

\begin{table}[!tb]
\renewcommand{\arraystretch}{1.05}
\setlength{\tabcolsep}{2pt}
\hspace{-2mm}
\scalebox{0.85}{
\begin{tabular}{l|cccc}
\Xhline{3\arrayrulewidth}
 Summarizer\vrule width 0pt height 11.5pt depth 5pt & 2-R & 2-F & 3-F  &  SU4-F \\\Xhline{3\arrayrulewidth}
 \hspace{-2mm} \scalebox{0.95}{\textbf{Trained on Our Corpus \scalebox{0.8}[0.85]{(size: 1000)}}} \vrule width 0pt height 12pt depth 5pt &   &  &   &   \\
 \scalebox{1}[1]{GCN} Hybrid 2 (Ours)\vrule width 0pt depth 0pt &  \textcolor{white}{{\bf *}}41.69{\bf *} & \textcolor{white}{{\bf *}}29.30{\bf *} & \textcolor{white}{{\bf *}}24.65{\bf *} & \textcolor{white}{{\bf *}}18.56{\bf *} \\
 \scalebox{1}[1]{GCN} Hybrid 1 (Ours)\vrule width 0pt height 0pt  & \textcolor{white}{{\bf *}}36.47{\bf *} & \textcolor{white}{{\bf *}}26.31{\bf *} & \textcolor{white}{{\bf *}}21.33{\bf *} & \textcolor{white}{{\bf *}}16.18{\bf *}\\
 \scalebox{1}[1]{GCN} Cited text spans (Ours) \vrule width 0pt height 0pt depth 5pt & \textcolor{white}{{\bf *}}{33.03}{\bf *} & \textcolor{white}{{\bf *}}{23.49}{\bf *} & \textcolor{white}{{\bf *}}{17.86}{\bf *} &  \textcolor{white}{{\bf *}}{14.15}{\bf *}
 \\\hdashline[1.5pt/2pt]
 \hspace{-1.5mm}\scalebox{0.95}{\textbf{Trained on CL-SciSumm \scalebox{0.8}[0.85]{(size: 30)}}} \vrule width 0pt height 12pt depth 5pt &   &  &   &   \\
 \scalebox{1}[1]{GCN} Cited text spans (Ours) ~ &  24.93& 18.46 & 12.77 & 12.21 \\
 Best participant 1
 \vrule width 0pt & 32.36 &21.94 & 16.79 & 13.63 \\
 Best participant 2
 \vrule width 0pt height 0pt depth 5pt & 26.67 &18.85 & 12.83  &  12.45\\
 \Xhline{3\arrayrulewidth}
 \multicolumn{4}{l}{\hspace{0mm}\scalebox{0.9}{{\bf *}: higher than all models trained on the CL-SciSumm corpus.} \vrule width 0pt height 12pt depth 0pt}\\[-2mm]
\end{tabular}
}
\caption{Results of {\bf Exp 1}, showing ROUGE evaluations on the CL-SciSumm Test benchmark. Models trained on our corpus outperform all the models trained on the existing CL-SciSumm Train set.
\vspace{-0mm}}
\label{tab:step1_result}
\end{table}

\begin{figure*}[!t]

    \setlength{\abovecaptionskip}{8pt}
    \centering
    \includegraphics[width=1.0\linewidth]{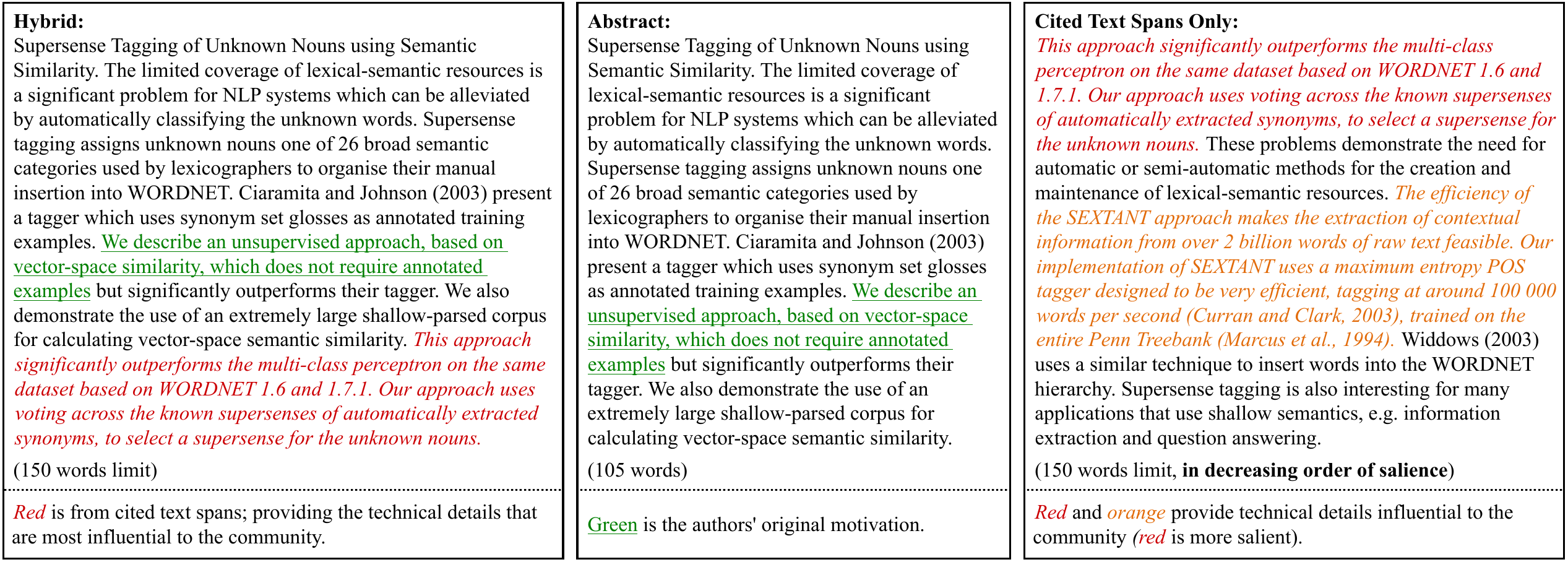}\vspace{-2.5mm}
    \caption{Comparison of our hybrid summary with the abstract and pure cited text spans summary, for paper P05-1004 in the CL-SciSumm 2016 test set.
    Our hybrid summary covers both the authors' original motivations (\uline{green}) and the technical details influential to the research community ({\it red}).\vspace{-2mm}}
    \label{fig:summ_comparison}
\end{figure*}

\begin{table}[!tb]
\renewcommand{\arraystretch}{1.05}
\centering
\scalebox{0.85}{
\begin{tabular}{l|cccc}
\Xhline{3\arrayrulewidth}
 Summarizer\vrule width 0pt height 11.5pt depth 5pt & 2-R & 2-F & 3-F &  SU4-F \\\Xhline{3\arrayrulewidth}
 Abstract\vrule width 0pt height 11.5pt depth 5pt & 29.52 & 29.40 & 23.16 &  23.34\\ \hline
 \scalebox{0.85}[0.9]{GCN} Hybrid 2 w/ auth\vrule width 0pt height 11.5pt depth 0pt & {\bf 33.88} & {\bf 31.54}& {\bf 24.32} & {\bf 24.36} \\
 \scalebox{0.85}[0.9]{GCN} Hybrid 2\vrule width 0pt height 0pt depth 0pt & 32.44 & 30.08 & 23.43 &  23.77\\
 \scalebox{0.85}[0.9]{GCN} Hybrid 1 w/ auth\vrule width 0pt height 0pt depth 0pt & 29.65 & 28.05 & 21.83 & 20.22 \\
 \scalebox{0.85}[0.9]{GCN} Hybrid 1\vrule width 0pt height 0pt depth 5pt & 29.64 & 27.96 & 21.81 & 19.41\\\hline
 \scalebox{0.85}[0.9]{GCN} Cited text spans w/ auth~\vrule width 0pt height 11.5pt depth 0pt & 26.30 & 24.39  & 18.85 & 17.31 \\
 \scalebox{0.85}[0.9]{GCN} Cited text spans\vrule width 0pt height 0pt depth 5pt & 25.16 & 24.26 & 18.79 & 17.67
 \\\Xhline{3\arrayrulewidth}
 \multicolumn{4}{l}{\hspace{0mm}\scalebox{0.9}{
 w/ auth: using authority feature.}\vrule width 0pt height 12pt depth 0pt}\\[-2mm]
\end{tabular}

}
\caption{Results of {\bf Exp 2}, showing ROUGE evaluations on the CL-SciSumm Test benchmark. All models are trained on our corpus. The hybrid models outperform abstracts and pure citation summaries.
\vspace{-2mm}}
\label{tab:step2_result}
\end{table}

\paragraph{Exp 1.}
Table \ref{tab:step1_result} shows the result of Exp 1, along with the top two participants in the CL-SciSumm shared task (\citeauthor{clscisumm_sys8,conroy2015vector_sys3}).
The upper part shows the model performance after training on our proposed corpus (1000 examples), and the lower part the existing CL-SciSumm corpus (30 examples).
We find that the neural model, {\it GCN Cited text spans},
performs on par with the participants when trained on CL-SciSumm, but when trained on our corpus, it gains significant boosts in all the ROUGE metrics (e.g., +5 in ROUGE-3-F) and greatly outperform all the models trained on CL-SciSumm.
With orders of magnitude more training examples than prior datasets, our corpus actually enables the data-driven neural network-based models to perform well on scientific paper summarization. This result suggests both the usefulness of the proposed corpus for training, and the feasibility of neural models in summarization given sufficient data.

\paragraph{Exp 2.}
Table \ref{tab:step2_result} shows the result of Exp 2, along with the baselines (Abstract and GCN Cited text spans).
We observe that both of the hybrid models perform clearly better than pure
cited text span summaries.
Moreover, Hybrid 1 surpasses abstracts in Recall, and
Hybrid 2 outperforms abstracts in all ROUGE metrics, including the F1 of R-2 and -3, which have the highest correlation with human judgments \cite{cohan2016revisiting}.
Hybrid 2 performs better than Hybrid 1, most likely
because Hybrid 2 builds on existing summaries (abstracts) and can ensure higher quality.
In this experiment, Hybrid 2 added two sentences on average to the original abstract.

To qualitatively study the advantage of the hybrid summarization,
we also compare and analyze the output summaries.
As an example, Figure \ref{fig:summ_comparison} shows the output summary of Hybrid 2 together with the abstract and pure cited text span summary for paper P05-1004 in the CL-SciSumm 2016 test set.
The hybrid summary, which augments the abstract
by taking in the most salient cited text spans, includes
the technical contributions that are most influential to the community but are not covered in the abstract ({\it red}).
The cited text span-based summary, on the other hand, provides more technical details, but
lacks some of the author's original messages such as the motivation and objective of their work (\uline{green}).
Thus, the hybrid summary is indeed more comprehensive than the abstract and cited text span summary because it incorporates both the authors' original insights and the 
community's views on the paper.

\textbf{Human evaluation.}
The above evaluation and analysis show the advantage of the hybrid models over the baselines.
Here we conduct human evaluation of the hybrid summaries against gold summaries to study their utility.
We asked 5 computer science students who took an NLP course to evaluate the coverage and coherence of the output summaries by our hybrid model, in a scale 1-5 (5 is the level of gold summaries).
The model achieved 4.5 and 4.2 on average for these two metrics. While there is room for improving coherence, these scores suggest that the model can generate comprehensive and readable summaries.

Finally, we observe in Table \ref{tab:step2_result} that all our models obtain moderate improvements by introducing the authority feature to reflect the authority of each citation made by the research community, suggesting the usefulness of this feature.


\section{Conclusion}
We proposed a novel dataset and hybrid models for scientific paper summarization.
Our corpus,
which contains
1,000 examples of papers, citation information and human summaries,
is orders of magnitude larger than prior datasets and facilitates future research in supervised scientific paper summarization.
We also presented hybrid summarization methods that integrate both authors' and
community's insights,
to
overcome the limitations of abstracts (may not convey actual impacts) and traditional citation-based summaries (may overlook authors' original messages).
Our experiments demonstrated that 1) the proposed dataset is indeed effective in training data-driven neural models,
and
that 2) the hybrid models produce more comprehensive summaries than abstracts and traditional citation-based summaries.
We hope that our large
annotated corpus and hybrid methods would open up new
avenues for scientific paper summarization.

\subsection*{Acknowledgements}
We thank Kokil Jaidka, Muthu Kumar Chandrasekaran, Min-Yen Kan, Yavuz Nuzumlali, Arman Cohan, as well as all the anonymous reviewers for their helpful feedback.
We also thank everyone who helped the evaluation in this work.

\bibliography{aaai19}

\begin{thebibliography}{}

\bibitem[\protect\citeauthoryear{Abu-Jbara and Radev}{2011}]{abuJbara&Radev11a}
Abu-Jbara, A., and Radev, D.~R.
\newblock 2011.
\newblock Coherent citation-based summarization of scientific papers.
\newblock In {\em ACL}.

\bibitem[\protect\citeauthoryear{Bergsma and
  Lin}{2006}]{bergsma-lin:2006:COLACL}
Bergsma, S., and Lin, D.
\newblock 2006.
\newblock Bootstrapping path-based pronoun resolution.
\newblock In {\em ACL}.

\bibitem[\protect\citeauthoryear{Cao \bgroup et al\mbox.\egroup
  }{2015}]{cao2015ranking}
Cao, Z.; Wei, F.; Dong, L.; Li, S.; and Zhou, M.
\newblock 2015.
\newblock Ranking with recursive neural networks and its application to
  multi-document summarization.
\newblock In {\em AAAI}.

\bibitem[\protect\citeauthoryear{Cao \bgroup et al\mbox.\egroup
  }{2017}]{cao2017improving}
Cao, Z.; Li, W.; Li, S.; and Wei, F.
\newblock 2017.
\newblock Improving multi-document summarization via text classification.
\newblock In {\em AAAI}.

\bibitem[\protect\citeauthoryear{Caruana, Lawrence, and
  Giles}{2001}]{caruana2001overfitting}
Caruana, R.; Lawrence, S.; and Giles, C.~L.
\newblock 2001.
\newblock Overfitting in neural nets: Backpropagation, conjugate gradient, and
  early stopping.
\newblock In {\em NIPS}.

\bibitem[\protect\citeauthoryear{Cheng and
  Lapata}{2016}]{cheng-lapata:2016:P16-1}
Cheng, J., and Lapata, M.
\newblock 2016.
\newblock Neural summarization by extracting sentences and words.
\newblock In {\em ACL}.

\bibitem[\protect\citeauthoryear{Cohan and Goharian}{2015}]{cohan2015sci}
Cohan, A., and Goharian, N.
\newblock 2015.
\newblock Scientific article summarization using citation-context and article's
  discourse structure.
\newblock In {\em EMNLP}.

\bibitem[\protect\citeauthoryear{Cohan and
  Goharian}{2016}]{cohan2016revisiting}
Cohan, A., and Goharian, N.
\newblock 2016.
\newblock Revisiting summarization evaluation for scientific articles.
\newblock In {\em LREC}.

\bibitem[\protect\citeauthoryear{Cohan and
  Goharian}{2017a}]{Cohan:2017:CCS:3077136.3080740}
Cohan, A., and Goharian, N.
\newblock 2017a.
\newblock Contextualizing citations for scientific summarization using word
  embeddings and domain knowledge.
\newblock In {\em SIGIR}.

\bibitem[\protect\citeauthoryear{Cohan and
  Goharian}{2017b}]{cohan2017scientific}
Cohan, A., and Goharian, N.
\newblock 2017b.
\newblock Scientific document summarization via citation contextualization and
  scientific discourse.
\newblock {\em IJDL}.

\bibitem[\protect\citeauthoryear{Cohan \bgroup et al\mbox.\egroup
  }{2018}]{cohan2018discourse}
Cohan, A.; Dernoncourt, F.; Kim, D.~S.; Bui, T.; Kim, S.; Chang, W.; and
  Goharian, N.
\newblock 2018.
\newblock A discourse-aware attention model for abstractive summarization of
  long documents.
\newblock In {\em NAACL-HLT}.

\bibitem[\protect\citeauthoryear{Cohen}{1960}]{cohen1960coefficient}
Cohen, J.
\newblock 1960.
\newblock A coefficient of agreement for nominal scales.
\newblock {\em Educational and psychological measurement}.

\bibitem[\protect\citeauthoryear{Collins, Augenstein, and
  Riedel}{2017}]{collins2017supervised}
Collins, E.; Augenstein, I.; and Riedel, S.
\newblock 2017.
\newblock A supervised approach to extractive summarisation of scientific
  papers.
\newblock In {\em CoNLL}.

\bibitem[\protect\citeauthoryear{Conroy and
  Davis}{2015}]{conroy2015vector_sys3}
Conroy, J., and Davis, S.
\newblock 2015.
\newblock Vector space and language models for scientific document
  summarization.
\newblock In {\em NAACL-HLT}.

\bibitem[\protect\citeauthoryear{Conroy and Davis}{2017}]{conroy2017section}
Conroy, J.~M., and Davis, S.~T.
\newblock 2017.
\newblock Section mixture models for scientific document summarization.
\newblock {\em IJDL}.

\bibitem[\protect\citeauthoryear{Elkiss \bgroup et al\mbox.\egroup
  }{2008}]{Elkiss:2008:BME:1331122.1331127}
Elkiss, A.; Shen, S.; Fader, A.; Erkan, G.; States, D.; and Radev, D.
\newblock 2008.
\newblock Blind men and elephants: What do citation summaries tell us about a
  research article?
\newblock {\em JASIST}.

\bibitem[\protect\citeauthoryear{Erkan and Radev}{2004}]{erkan2004lexrank}
Erkan, G., and Radev, D.~R.
\newblock 2004.
\newblock Lexrank: Graph-based lexical centrality as salience in text
  summarization.
\newblock {\em JAIR}.

\bibitem[\protect\citeauthoryear{Hochreiter and
  Schmidhuber}{1997}]{Hochreiter:1997:LSM:1246443.1246450}
Hochreiter, S., and Schmidhuber, J.
\newblock 1997.
\newblock Long short-term memory.
\newblock {\em Neural Computation} 9(8):1735--1780.

\bibitem[\protect\citeauthoryear{Hong and
  Nenkova}{2014}]{hong-nenkova:2014:EACL}
Hong, K., and Nenkova, A.
\newblock 2014.
\newblock Improving the estimation of word importance for news multi-document
  summarization.
\newblock In {\em EACL}.

\bibitem[\protect\citeauthoryear{Jaidka \bgroup et al\mbox.\egroup
  }{2016}]{jaidka2016overview}
Jaidka, K.; Chandrasekaran, M.~K.; Rustagi, S.; and Kan, M.-Y.
\newblock 2016.
\newblock Overview of the cl-scisumm 2016 shared task.
\newblock In {\em BIRNDL}.

\bibitem[\protect\citeauthoryear{Jaidka \bgroup et al\mbox.\egroup
  }{2017}]{jaidka2017overview}
Jaidka, K.; Chandrasekaran, M.~K.; Jain, D.; and Kan, M.-Y.
\newblock 2017.
\newblock The cl-scisumm shared task 2017: Results and key insights.
\newblock In {\em BIRNDL}.

\bibitem[\protect\citeauthoryear{Jaidka \bgroup et al\mbox.\egroup
  }{2018}]{jaidka2018overview}
Jaidka, K.; Yasunaga, M.; Chandrasekaran, M.~K.; Radev, D.; and Kan, M.-Y.
\newblock 2018.
\newblock The cl-scisumm shared task 2018: Results and key insights.
\newblock In {\em BIRNDL @SIGIR}.

\bibitem[\protect\citeauthoryear{Kingma and Ba}{2015}]{kingma2015adam}
Kingma, D., and Ba, J.
\newblock 2015.
\newblock Adam: A method for stochastic optimization.
\newblock In {\em ICLR}.

\bibitem[\protect\citeauthoryear{Kipf and Welling}{2017}]{kipf2017semi}
Kipf, T., and Welling, M.
\newblock 2017.
\newblock Semi-supervised classification with graph convolutional networks.
\newblock In {\em ICLR}.

\bibitem[\protect\citeauthoryear{Landis and Koch}{1977}]{landis1977measurement}
Landis, J.~R., and Koch, G.~G.
\newblock 1977.
\newblock The measurement of observer agreement for categorical data.
\newblock {\em Biometrics}.

\bibitem[\protect\citeauthoryear{Li \bgroup et al\mbox.\egroup
  }{2016}]{clscisumm_sys8}
Li, L.; Mao, L.; Zhang, Y.; Chi, J.; Huang, T.; Cong, X.; and Peng, H.
\newblock 2016.
\newblock Cist system for cl-scisumm 2016 shared task.
\newblock In {\em BIRNDL}.

\bibitem[\protect\citeauthoryear{Li \bgroup et al\mbox.\egroup
  }{2017}]{li2017computational}
Li, L.; Mao, L.; Zhang, Y.; Chi, J.; Huang, T.; Cong, X.; and Peng, H.
\newblock 2017.
\newblock Computational linguistics literature and citations oriented citation
  linkage, classification and summarization.
\newblock {\em IJDL}.

\bibitem[\protect\citeauthoryear{Lin}{2004}]{lin2004rouge}
Lin, C.-Y.
\newblock 2004.
\newblock Rouge: A package for automatic evaluation of summaries.
\newblock In {\em Text summarization branches out: Proceedings of the ACL-04
  workshop}.

\bibitem[\protect\citeauthoryear{Lloret, Rom{\'a}-Ferri, and
  Palomar}{2013}]{lloret2013compendium}
Lloret, E.; Rom{\'a}-Ferri, M.~T.; and Palomar, M.
\newblock 2013.
\newblock Compendium: A text summarization system for generating abstracts of
  research papers.
\newblock {\em Data \& Knowledge Engineering}.

\bibitem[\protect\citeauthoryear{Luhn}{1958}]{luhn1958automatic}
Luhn, H.~P.
\newblock 1958.
\newblock The automatic creation of literature abstracts.
\newblock {\em IBM Journal of research and development}.

\bibitem[\protect\citeauthoryear{Marcheggiani and
  Titov}{2017}]{marcheggiani2017encoding}
Marcheggiani, D., and Titov, I.
\newblock 2017.
\newblock Encoding sentences with graph convolutional networks for semantic
  role labeling.
\newblock In {\em EMNLP}.

\bibitem[\protect\citeauthoryear{Mei and Zhai}{2008}]{mei-zhai:2008:ACLMain}
Mei, Q., and Zhai, C.
\newblock 2008.
\newblock Generating impact-based summaries for scientific literature.
\newblock In {\em ACL-08: HLT}.

\bibitem[\protect\citeauthoryear{Nakov, Schwartz, and
  Hearst}{2004}]{nakov2004citances}
Nakov, P.~I.; Schwartz, A.~S.; and Hearst, M.
\newblock 2004.
\newblock Citances: Citation sentences for semantic analysis of bioscience
  text.
\newblock In {\em SIGIR}.

\bibitem[\protect\citeauthoryear{Nallapati, Zhai, and
  Zhou}{2017}]{nallapati2017summarunner}
Nallapati, R.; Zhai, F.; and Zhou, B.
\newblock 2017.
\newblock Summarunner: A recurrent neural network based sequence model for
  extractive summarization of documents.
\newblock In {\em AAAI}.

\bibitem[\protect\citeauthoryear{Narayan, Cohen, and
  Lapata}{2018}]{narayan2018ranking}
Narayan, S.; Cohen, S.~B.; and Lapata, M.
\newblock 2018.
\newblock Ranking sentences for extractive summarization with reinforcement
  learning.
\newblock In {\em NAACL}.

\bibitem[\protect\citeauthoryear{Paice and
  Jones}{1993}]{Paice:1993:IIC:160688.160696}
Paice, C.~D., and Jones, P.~A.
\newblock 1993.
\newblock The identification of important concepts in highly structured
  technical papers.
\newblock In {\em SIGIR}.

\bibitem[\protect\citeauthoryear{Paice}{1981}]{Paice:1980:AGL:636669.636680}
Paice, C.~D.
\newblock 1981.
\newblock The automatic generation of literature abstracts: An approach based
  on the identification of self-indicating phrases.
\newblock In {\em SIGIR}.

\bibitem[\protect\citeauthoryear{Parveen, Mesgar, and
  Strube}{2016}]{parveen2016generating}
Parveen, D.; Mesgar, M.; and Strube, M.
\newblock 2016.
\newblock Generating coherent summaries of scientific articles using coherence
  patterns.
\newblock In {\em EMNLP}.

\bibitem[\protect\citeauthoryear{Parveen, Ramsl, and
  Strube}{2015}]{parveen-ramsl-strube:2015:EMNLP}
Parveen, D.; Ramsl, H.-M.; and Strube, M.
\newblock 2015.
\newblock Topical coherence for graph-based extractive summarization.
\newblock In {\em EMNLP}.

\bibitem[\protect\citeauthoryear{Pascanu, Mikolov, and
  Bengio}{2012}]{Pascanu2012}
Pascanu, R.; Mikolov, T.; and Bengio, Y.
\newblock 2012.
\newblock On the difficulty of training recurrent neural networks.
\newblock {\em arXiv preprint arXiv:1211.5063}.

\bibitem[\protect\citeauthoryear{Pennington, Socher, and
  Manning}{2014}]{pennington-socher-manning:2014:EMNLP2014}
Pennington, J.; Socher, R.; and Manning, C.
\newblock 2014.
\newblock Glove: Global vectors for word representation.
\newblock In {\em EMNLP}.

\bibitem[\protect\citeauthoryear{Qazvinian and
  Radev}{2008}]{qazvinian2008scientific}
Qazvinian, V., and Radev, D.~R.
\newblock 2008.
\newblock Scientific paper summarization using citation summary networks.
\newblock In {\em COLING}.

\bibitem[\protect\citeauthoryear{Radev \bgroup et al\mbox.\egroup
  }{2013}]{aan_2013}
Radev, D.~R.; Muthukrishnan, P.; Qazvinian, V.; and Abu-Jbara, A.
\newblock 2013.
\newblock {The ACL anthology network corpus}.
\newblock {\em Language Resources and Evaluation}  1--26.

\bibitem[\protect\citeauthoryear{See, Liu, and Manning}{2017}]{see2017get}
See, A.; Liu, P.; and Manning, C.
\newblock 2017.
\newblock Get to the point: Summarization with pointer-generator networks.
\newblock In {\em ACL}.

\bibitem[\protect\citeauthoryear{Siddharthan and
  Teufel}{2007}]{siddharthan2007whose}
Siddharthan, A., and Teufel, S.
\newblock 2007.
\newblock Whose idea was this, and why does it matter? attributing scientific
  work to citations.
\newblock In {\em HLT-NAACL}.

\bibitem[\protect\citeauthoryear{Srivastava \bgroup et al\mbox.\egroup
  }{2014}]{JMLR:v15:srivastava14a}
Srivastava, N.; Hinton, G.; Krizhevsky, A.; Sutskever, I.; and Salakhutdinov,
  R.
\newblock 2014.
\newblock Dropout: A simple way to prevent neural networks from overfitting.
\newblock {\em JMLR} 15:1929--1958.

\bibitem[\protect\citeauthoryear{Teufel and
  Moens}{2002}]{Teufel:2002:SSA:638178.638180}
Teufel, S., and Moens, M.
\newblock 2002.
\newblock Summarizing scientific articles: Experiments with relevance and
  rhetorical status.
\newblock {\em Computational linguistics} 28(4):409--445.

\bibitem[\protect\citeauthoryear{Woodsend and
  Lapata}{2010}]{woodsend-lapata:2010:ACL}
Woodsend, K., and Lapata, M.
\newblock 2010.
\newblock Automatic generation of story highlights.
\newblock In {\em ACL}.

\bibitem[\protect\citeauthoryear{Yasunaga \bgroup et al\mbox.\egroup
  }{2017}]{yasunagaetal2017}
Yasunaga, M.; Zhang, R.; Meelu, K.; Pareek, A.; Srinivasan, K.; and Radev,
  D.~R.
\newblock 2017.
\newblock Graph-based neural multi-document summarization.
\newblock In {\em CoNLL-2017}.

\end{thebibliography}
\bibliographystyle{aaai19}

\end{document}